\pdfoutput=1

\documentclass[11pt]{article}

\usepackage[final]{paper}

\usepackage{times}
\usepackage{latexsym}

\usepackage{graphicx}
\usepackage{dblfloatfix}
\usepackage{algorithm}
\usepackage{algorithmic}

\usepackage{amssymb}
\usepackage{amsmath}
\usepackage{cleveref}
\usepackage{listings}

\usepackage[T1]{fontenc}
\usepackage[utf8]{inputenc}
\usepackage{microtype}
\usepackage{inconsolata}

\title{Reinforcement learning for question answering in programming domain using public community scoring as a human feedback}

\author{Alexey Gorbatovski \\
  ITMO University \\
  Saint Petersburg, Russia \\
  \texttt{gorbatoski@itmo.ru} \\\And
  Sergey Kovalchuk \\
  Huawei \\
  Saint Petersburg, Russia \\
  \texttt{sergey.kovalchuk@huawei.com} \\}

\begin{document}
\maketitle
\begin{abstract}

In this study, we investigate the enhancement of the GPT Neo 125M's performance in Community Question Answering (CQA) with a focus on programming, through the integration of Reinforcement Learning from Human Feedback (RLHF) and the utilization of scores from Stack Overflow. Two distinct reward model training strategies are employed for fine-tuning with Proximal Policy Optimization (PPO). Notably, the improvements in performance achieved through this method are comparable to those of GPT Neo’s 2.7B parameter variant. Additionally, an auxiliary scoring mechanism is introduced, which demonstrates the limitations of conventional linguistic metrics in evaluating responses in the programming domain. Through accurate analysis, this paper looks at the divergence between traditional linguistic metrics and our human-preferences-based reward model, underscoring the imperative for domain-specific evaluation methods. By elucidating the complexities involved in applying RLHF to programming CQA and accentuating the significance of context-aware evaluation, this study contributes to the ongoing efforts in refining Large Language Models through focused human feedback.

\end{abstract}

\section{Introduction}

Advances in Reinforcement Learning from Human Feedback (RLHF) have revolutionized the fine-tuning of Large Language Models (LLMs), facilitating adaptation for human-like response generation and precise behavior control \citep{ouyang2022training}. While RLHF has proven effective in general domains, its application in specialized fields such as Community Question Answering (CQA) for programming remains unexplored \citep{beeching2023stackllama}. LLMs face unique challenges in handling the complex nature of programming queries, including conceptual understanding, code generation, API usage, and debugging, due to struggles with subtle semantic relations.

Furthermore, a critical challenge is the evaluation of the quality of responses generated by LLMs. Conventional metrics such as BertScore and Rouge do not capture the essence of responses effectively, especially in specialized domains like programming \citep{wang2019text}. Moreover, they don’t account for the diversity in valid answers and lack in capturing deeper semantic correctness. The development of more reliable and context-sensitive evaluation metrics is essential\citep{kovalchuk-etal-2022-human}.

To address these challenges, in this paper, we investigate the application of RLHF to a smaller model, GPT Neo 125M \citep{black2021gpt}, in the context of programming CQA. We aim not only to enhance the model's response generation capabilities but also to address the evaluation challenge. Our contributions are two-fold. First, we explore the potential and efficacy of RLHF in retraining a smaller LLM for programming CQA. Second, through empirical analysis, we highlight the discrepancies between the RLHF reward model and existing linguistic metrics, emphasizing the limitations of current evaluation methodologies and advocating for the development of more semantically-sensitive measures.

The structure of the paper is as follows: Section 2 provides background information and describes the datasets used in this study. In Section 3, we delve into the application of RLHF for programming CQA, explaining the data preparing methodologies employed. Section 4 focuses on the experimental evaluation and results. Section 5 presents a discussion of the study results and evaluation methods. Finally, Section 6 concludes the paper with final remarks and reflections on our findings.

\section{Background and Dataset}

\subsection{Background on RLHF and LLMs}

Reinforcement Learning from Human Feedback (RLHF) is a technique where models are trained using human feedback as rewards. This method has become notably beneficial in refining the performance and behavior control of Large Language Models (LLMs). RLHF initiates with models trained using supervised fine-tuning (SFT), which are then iteratively improved. Crucially, the human scoring process in RLHF is often automated by training a separate reward model, serving as an optimization proxy. This process varies in implementation and warrants further exploration.

The application of RLHF in LLMs has been explored in various contexts. For instance, \citet{ziegler2019fine} studied the impact of reward learning on specific tasks, demonstrating the potential of RLHF in enhancing the performance of LLMs. The work of \citet{stiennon2020learning} and the OpenAI Alignment Team in 2021 further expanded the scope of RLHF, applying it to the task of summarizing text and books, respectively.

In the context of Question Answering (QA), RLHF has been used to train models to navigate the web \citep{nakano2021webgpt} and to follow instructions \citep{ouyang2022training}. However, these studies have mainly focused on general domains or specific tasks, and the application of RLHF in specialized fields such as programming Community Question Answering (CQA) remains largely unexplored.

\subsection{Dataset Selection and Preprocessing}

In the study, we used Stack Overflow\footnote{\url{https://stackoverflow.com/}} (SO) as the primary data source for programming-related question-answering tasks. We regarded the answers on SO as reference solutions. We compiled the dataset from the original data available on Stack Exchange\footnote{\url{https://stackexchange.com/}}, focusing on questions specifically tagged with 'python'. This dataset, which includes titles, questions, answers per question, and user scores for each, was used for both supervised fine-tuning and partial reward model training. To ensure the dataset's relevance and homogeneity, we subjected it to a series of constraints and transformations. Furthermore, we adjusted user ratings for different reward model training setups.

We applied several constraints to refine the dataset and maintain consistency:

\begin{itemize}
    \item We only selected questions classified as “API Usage” according to the taxonomy by \citet{beyer2020kind}. This selection was performed using regular expressions to ensure alignment with the study's focus.
    \item To maintain text purity and the ability to generate a response based on the context only, we filtered out questions and answers containing images, links, or code blocks, designated by the <pre><code> HTML tags. Code blocks were filtered out to prevent the model from generating code snippets during training, as this would significantly complicate the evaluation process due to the lack of established metrics for assessing the quality of generated code.
    \item HTML content was sanitized and converted to plain text using the Beautiful Soup\footnote{\url{https://github.com/wention/BeautifulSoup4}} library to prepare it for natural language processing.
\end{itemize}

We obtained a dataset of 6,946 training entries and 1,000 validation entries. To prevent data leakage and ensure temporal relevance, the validation set included questions posted after December 14, 2021.

\begin{figure*}[t]
\centering
\includegraphics[width=0.8\textwidth]{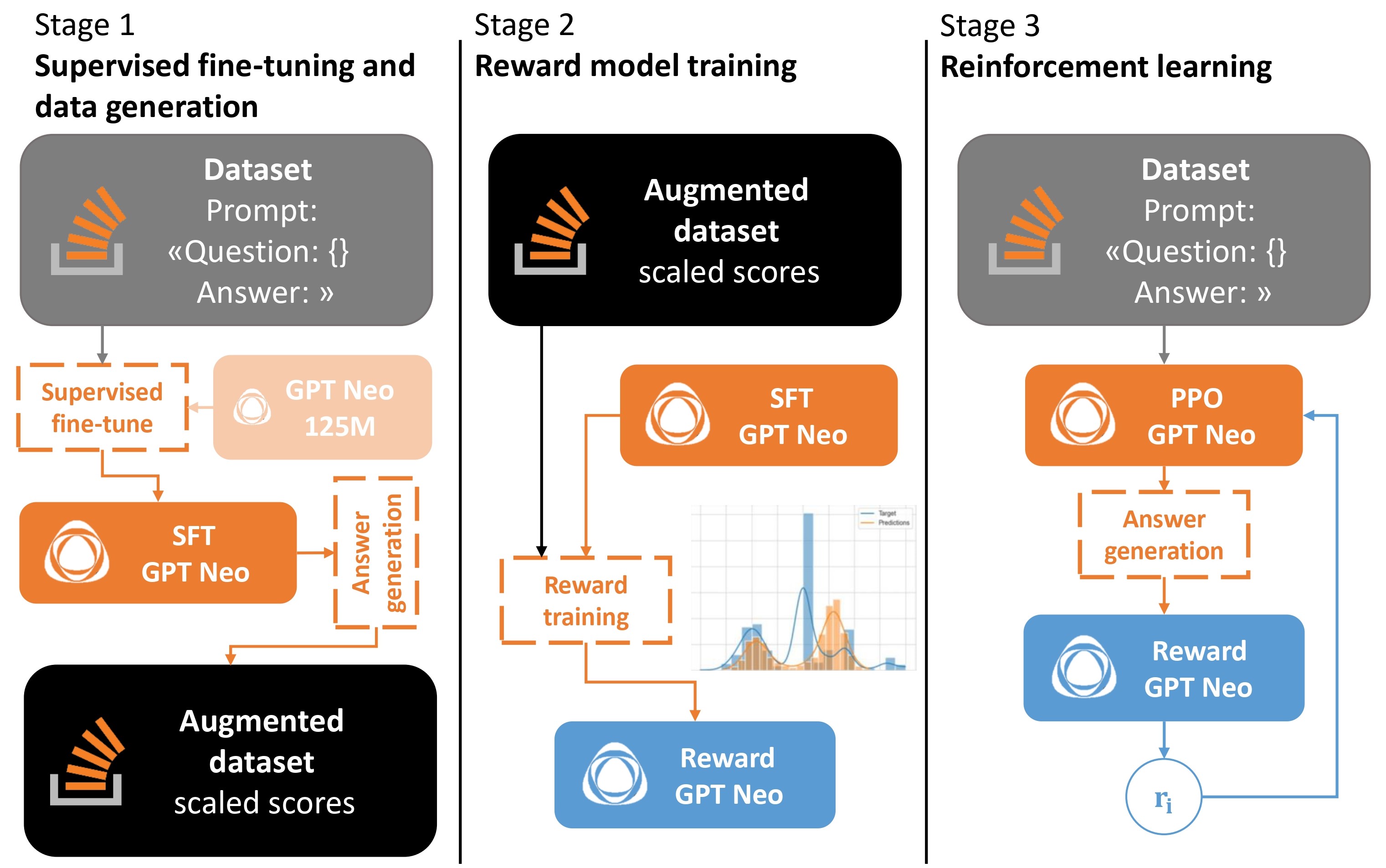}
\caption{The general schema of Reinforcement Learning from Human Feedback for programming Q\&A}
\label{fig:rlhf_schema}
\end{figure*}

While this dataset is highly relevant for studying RLHF in programming CQA, it is worth noting that the constraints applied may introduce certain limitations in terms of the diversity of questions and real-world applicability.

\section{RLHF for programming Q\&A}

The general schema of RLHF utilized in this study consists of several stages, as depicted in Fig. \ref{fig:rlhf_schema}. The process commences with the training of an initial policy via supervised learning. Subsequently, a reward model is trained to acquire human preferences from the labeled data. Finally, the policy is fine-tuned using Proximal Policy Optimization (PPO) \citep{schulman2017proximal}, guided by the reward model.

In this study, we have adapted RLHF for programming Q\&A by converting user ratings from Stack Overflow into feedback for training our model. We used two distinct approaches: creating regression scores and contrastive scores for the straightforward comparison of answers. Additionally, to enhance the logical alignment of sentences and mitigate the model's errors in generation, we completed the dataset for reward model training with generations from the SFT model.

\subsection{Transformation of User Ratings}

To account for biases arising from factors like the question's age and popularity, we preprocess and normalize user ratings. Our approach comprises two distinct transformations: Regression Scores and Contrastive Scores.

\begin{algorithm}
\caption{Regression scores transforming}
\label{algo:regression_scores}
\begin{algorithmic}[1]
\REQUIRE{user votes for each answer $N\_votes_{ij}$}
\ENSURE{regression scores $s_{ij}$}
\FOR{each question $q_i$}
    \FOR{each answer $a_j$ in $q_i$}
        \STATE $s_{ij} = \frac{N\_votes_{ij}}{N\_answers_i}$
    \ENDFOR
\ENDFOR
\STATE $l\_bound, u\_bound = 1.5 \times IQR(s_{ij})$
\FOR{each score $s_{ij}$}
    \IF{$s_{ij}$ outside the range $[l\_bound, u\_bound]$}
        \STATE $s_{ij} = clip(s_{ij}, -1, 1)$
    \ELSE
        \STATE $s_{ij} = max\_abs\_scale(s_{ij}, sign(s_{ij}))$
    \ENDIF
\ENDFOR
\RETURN $s_{ij}$ for all $i, j$
\end{algorithmic}
\end{algorithm}

\subsubsection{Regression Scores}

For regression scores, user ratings were normalized by the total number of answers for each question. After clipping outliers, the ratings were scaled to control the standard deviation, thus stabilizing the regression training. The process is outlined in Algorithm~\ref{algo:regression_scores}. There $v_{ij}$ represents the votes for each answer $a_j$ in question $q_i$ and $s_{ij}$ for the regression scores.

\subsubsection{Contrastive Scores}

Contrastive scores allow a convenient comparison of answer ratings by mapping them to logarithmically scaled scores \citep{askell2021general}. Accepted answers receive an additional increment, while negative ratings are assigned a standard value. The following Algorithm~\ref{algo:contrasted_scores} details this process:

\begin{algorithm}
\caption{Contrastive scores scaling}
\label{algo:contrasted_scores}
\begin{algorithmic}[1]
\REQUIRE{votes for each answer $v_j$}
\ENSURE{contrastive scores $s_ij$}
\FOR{each answer $a_j$}
    \IF{$v_j < 0$}
        \STATE $s_j = -1$
    \ELSE
        \STATE $s_j = \lceil log_2(1 + v_j) \rceil$
        \IF{$a_j$ is accepted}
            \STATE $s_j = s_j + 1$
        \ENDIF
    \ENDIF
\ENDFOR
\FOR{each question $q_i$}
    \IF{$N\_answers_i > 1$}
        \STATE $s_{max} = max(s_{ij})$ for all $j$
        \FOR{each answer $a_k$ in $q_i$}
        \STATE compare score $s_k$ with $s_{max}$
    \ENDFOR
    \ENDIF
\ENDFOR
\RETURN compared pairs $\{(a_j, a_k)\}$
\end{algorithmic}
\end{algorithm}

In this algorithm, $v_j$ denotes the votes for each answer, and the contrastive scores, $s_j$, are computed using a logarithmic scale.

Additionally, we first identified questions with more than one answer after preprocessing and filtering, which amounted to 3,076. We compared each of these answers with the highest-voted answer for each question during the training of the reward model. After this comparison, the contrast dataset contained 1,804 rows.

\subsection{Data Generation for Reward Model Training}

We generated 6,872 additional answers for questions with only one answer to create a comparison set essential for training the reward model. This step was undertaken to ensure a diverse dataset that simulates various answer qualities.

For the regression approach, we assigned these generated answers a normal distribution $\mathcal{N}(-0.5, 0.1^2)$. This was based on the observation that most generated answers were either completely uninformative or erroneous. We believe that discouraging the generation of nonsensical answers is a helpful practice.

In the contrastive approach, these generated answers were incorporated into the contrast dataset, which previously only included questions with more than one existing answer.

The generation of this additional data was crucial for robustly training the reward model. In the experimental section, we will delve into how this dataset was leveraged to train the reward model effectively, and the evaluation metrics used to assess its performance.

\section{Experimental evaluation}

\begin{figure*}[h]
\centering
\includegraphics[width=0.8\textwidth]{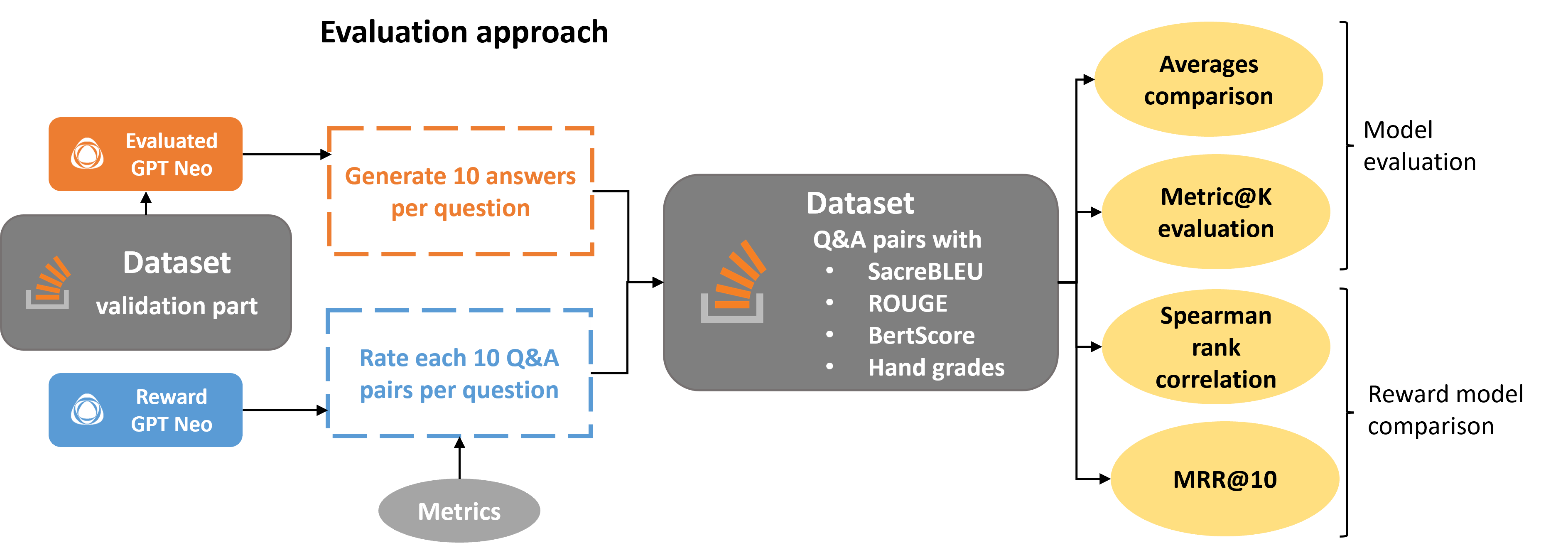}
\caption{The general schema of evaluation approach}
\label{fig:eval_schema}
\end{figure*}

This section aims to evaluate the effectiveness of the RLHF training approach for improving the quality of generated responses in the programming QA domain. Specifically, we compare three versions of the model, according to Fig. \ref{fig:rlhf_schema} - the base model, the SFT version, and the RLHF version. The evaluation focuses on the performance of the reward model training methods and the generated responses' quality.

\subsection{Evaluation approach}

Fig. \ref{fig:eval_schema} illustrates the evaluation schema. For each question in the validation dataset, the model generates ten responses using sampling-based decoding. This approach allows us to study the average quality of the generated responses without bias toward the worst or best cases. The parameters used for sampling-based decoding are as follows:

\begin{itemize}
    \item \texttt{do\_sample}: true
    \item \texttt{no\_repeat\_ngram\_size}: 2
    \item \texttt{top\_k}: 50
    \item \texttt{top\_p}: 0.9
\end{itemize}

To evaluate the responses' content and semantic similarity, we employ the SacreBLEU, Rouge, and BertScore metrics as common metrics for natural language generation tasks. Additionally, the reward models rate each generated response as an alternative quality assessment tool.

For a more insightful evaluation, we also conduct a human-based assessment. A subset of 100 randomly selected questions is manually evaluated by ourselves. Each generated answer for these questions is inspected and marked as useful (1) or not useful (0) for solving the problem stated in the question. This binary labeling enables the computation of the Mean Reciprocal Rank (MRR), which assesses the relevance of the generated responses.

Finally, to investigate the consistency between the different metrics and reward model assessments, we employ Spearman’s rank correlation coefficient. This statistical measure will provide insight into whether the automatic metrics and the reward model assessments are aligned in evaluating response quality.

\subsection{Experimental setup}

All experiments were conducted using the GPT Neo model \citep{gpt-neo} with 125 million parameters, selected based on the constraints discussed in the corresponding section.

\subsubsection{Supervised fine-tuning}

Fine-tuning was performed on the training dataset described in Section 2.2. We utilized the Transformers and PyTorch Lightning libraries with the following hyperparameters: optimizer = Adam, Adam betas = (0.9, 0.999), Adam epsilon = 1e-08, weight decay = 0.001, learning rate = 2e-05, learning rate decay scheme = linear, batch size = 12, and mixed precision training (fp16). The maximum length of the concatenated question and answer was set to 512 tokens. If the input sequence exceeded this length, the question was truncated, ensuring that the full answer was available for training.

\subsubsection{Reward Model Training}

We trained the reward model using two approaches: regression and answer comparison. The regression approach employed the Mean Squared Error (MSE) as the loss function, while the answer comparison approach used the Contrastive Loss (see Formula~\ref{eq:contrastive_loss}).

\begin{equation}
\label{eq:contrastive_loss}
\small
\mathcal{L}(\theta) = -\mathbb{E}_{(x,y_j,y_k)\sim D}[\log(\sigma(r_\theta(x,y_j) - r_\theta(x,y_k)))]
\end{equation}

where $r$ and $y$ are the reward model’s score and $y$ is the preferred candidate respectively. Both approaches used the SFT model as the basis.

For the regression approach, the hyperparameters were as follows: optimizer = Adam, Adam betas = (0.9, 0.999), Adam epsilon = 1e-08, weight decay = 0.001, learning rate = 2e-05, learning rate decay scheme = linear, batch size = 16, and mixed precision training (fp16).

For the contrastive approach, we used the same hyperparameters with a different learning rate (3e-05) and batch size (8). Additionally, for both approaches, the weights of the first and last layers, as well as all linear layers of the model, were updated during training.

Both approaches exhibited stability during training, achieving validation accuracies of 93\% and 95\% respectively. For the regression approach, accuracy was computed using the formula:
\begin{equation}
\small
accuracy = \frac{1}{n}\sum[sign(r_i) = y_i]
\end{equation}
where $r_i$ and $y_i$ mean $i^{th}$ reward and target score respectively. Considering both positive and negative rewards is essential in reinforcement learning.

\subsubsection{Fine-tuning with RL}

We employed reinforcement learning using the TRL and Transformers libraries, with typical RLHF parameters: optimizer = Adam, Adam betas = (0.9, 0.95), Adam epsilon = 1e-08, learning rate = 1.41e-05, epsilon clip range = 0.2, buffer size = 64, and batch size = 16. Additionally, we used adaptive KL control with an initial KL coefficient of 0.2 and a target of 6.

During reinforcement learning, the training was stable, and the average reward increased when using the reward model based on the regression approach. However, training was unstable and did not converge using the reward model based on answer comparisons. Tuning model starts generate repetitive words and incoherent sentences. Therefore, the results section presents the outcomes for the model trained using the regression-based reward model.

\begin{table*}
\centering
\begin{tabular}{l@{\hspace{3mm}}c@{\hspace{3mm}}c@{\hspace{3mm}}c@{\hspace{3mm}}c@{\hspace{3mm}}c@{\hspace{3mm}}c}
\hline
 & \textbf{SacreBLEU} & \textbf{Rouge 1} & \textbf{Rouge 2} & \textbf{BertScore} & \textbf{Reg. Reward} & \textbf{Contr. Reward} \\
\hline
Base 125M & 0.0433 & 0.1816 & 0.0233 & 0.9420 & -0.1479 & -1.0124 \\
 & {\small $(\sigma: 0.0071)$} & {\small $(\sigma: 0.0684)$} & {\small $(\sigma: 0.0160)$} & {\small $(\sigma: 0.0057)$} & {\small $(\sigma: 0.0994)$} & {\small $(\sigma: 1.0214)$} \\

SFT 125M & 0.0484 & 0.1903 & 0.0237 & 0.9483 & 0.1257 & -0.0173 \\
& {\small $(\sigma: 0.0088)$} & {\small $(\sigma: 0.0581)$} & {\small $(\sigma: 0.0151)$} & {\small $(\sigma: 0.0097)$} & {\small $(\sigma: 0.0864)$} & {\small $(\sigma: 1.0123)$} \\

RLHF 125M & \textbf{0.0489} & 0.1884 & 0.0230 & \textbf{0.9493} & \textbf{0.1869} & \textbf{0.3955} \\
& {\small $(\sigma: 0.0092)$} & {\small $(\sigma: 0.0545)$} & {\small $(\sigma: 0.0149)$} & {\small $(\sigma: 0.0105)$} & {\small $(\sigma: 0.0767)$} & {\small $(\sigma: 0.9720)$} \\

Base 2.7B & 0.0455 & \textbf{0.1906} & \textbf{0.0275} & 0.9417 & -0.1123 & -0.0365 \\
& {\small $(\sigma: 0.0073)$} & {\small $(\sigma: 0.0735)$} & {\small $(\sigma: 0.0190)$} & {\small $(\sigma: 0.0054)$} & {\small $(\sigma: 0.1045)$} & {\small $(\sigma: 1.1245)$} \\
\hline
\end{tabular}
\caption{\label{table:averages}
Comparison of average metrics for different models. Each entry contains the mean of the corresponding metric across ten generation attempts.
}
\end{table*}

\subsection{Results}

This section presents the results of the experiments, which were conducted to assess the efficacy of the RLHF training approach in the context of programming QA response generation. We examine the performance of the different models and discuss the correlation and consistency between the metrics employed for evaluating the quality of the generated responses.

\subsubsection{Comparison of Average Metrics}

Our evaluation process involved computing the average metrics for ten generation attempts by four models: Base GPT Neo 125M (Base 125M), Supervised Fine-tuning GPT Neo 125M (SFT 125M), RLHF GPT Neo 125M (RLHF 125M), and Base GPT Neo 2.7B (Base 2.7B). These models evaluated using several metrics, including SacreBLEU, Rouge 1, Rouge 2 and BertScore, as well as the scores obtained from the regression and contrastive reward models.

Table~\ref{table:averages} presents the average values of these metrics for each model. Notably, the RLHF version demonstrated superior performance compared to the SFT model in terms of SacreBLEU and BertScore. However, the larger Base GPT Neo 2.7B model surpassed the other models in terms of the Rouge scores. All highlighted metrics were deemed statistically significant via the KS-test. The inclusion of bootstrapped confidence intervals further clarifies the model's improvement relative to the baseline.

\subsection{Metrics at Rank k Analysis}

Beyond the mean values, we performed an in-depth analysis using the metric@k approach. The term “metric at rank k” refers to the highest score achieved by a metric among k randomly sampled generation attempts. This analysis helps to reveal the capability of the models to generate high-quality responses within a certain number of attempts.

\begin{figure*}[h]
\centering
\includegraphics[width=0.9\textwidth]{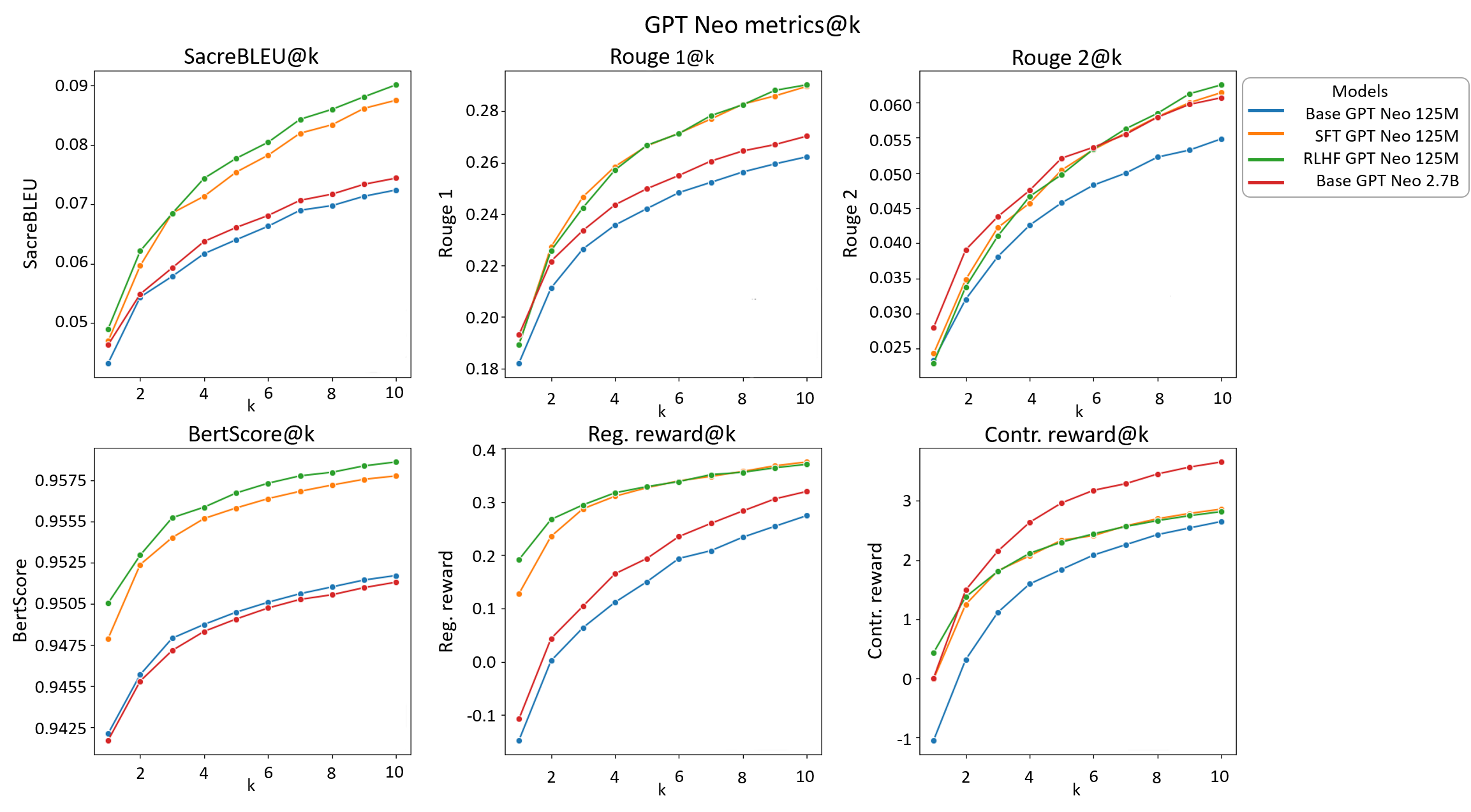}
\caption{Graphs of dependencies of metric values on the number of k attempts to generate}
\label{fig:metric_k}
\end{figure*}

Fig. ~\ref{fig:metric_k} illustrates graphs depicting the relationship between the metric values and the number of generation attempts (k). These graphs provide insights into the performance of the models as the number of generation attempts increases. Particularly, after several generation attempts, both the SFT and RLHF versions appear to outperform the larger GPT Neo 2.7B model in terms of the evaluated metrics. Additionally, the RLHF model exhibits significant improvement in BertScore, which suggests enhanced semantic similarity between the generated and reference responses.

\begin{table*}
\centering
\begin{tabular}{l@{\hspace{3mm}}c@{\hspace{3mm}}c@{\hspace{3mm}}c@{\hspace{3mm}}c@{\hspace{3mm}}c@{\hspace{3mm}}c}
\hline
 & \textbf{SacreBLEU} & \textbf{Rouge 1} & \textbf{Rouge 2} & \textbf{BertScore} & \textbf{Reg. Reward} & \textbf{Contr. Reward} \\
\hline
Base 125M & 0.0724 & 0.2623 & 0.0549 & 0.9517 & 0.2746 & 2.6521 \\
SFT 125M & 0.0875 & 0.2897 & 0.0614 & 0.9578 & \textbf{0.3754} & 2.8612 \\
RLHF 125M & \textbf{0.0901} & \textbf{0.2903} & \textbf{0.0625} & \textbf{0.9586} & 0.3711 & 2.8187 \\
Base 2.7B & 0.0744 & 0.2704 & 0.0607 & 0.9513 & 0.3201 & \textbf{3.6581} \\
\hline
\end{tabular}
\caption{\label{table:metrics_at_rank_10}Comparison of average metrics@10 for different models.}
\end{table*}

Table~\ref{table:metrics_at_rank_10} presents the results for Metrics@10, which indicates the best metric scores among 10 generation attempts.  Most of the highlighted values have been determined to be statistically significant as per the U-test and KS-test. Interestingly, Base GPT Neo 2.7B exhibits the highest average model reward based on the contrastive approach. This might suggest that the model’s responses are more diverse and, in some cases, closer to the reference answers.

The Rouge 2 metric, which focuses on the overlap of bigrams between the generated and reference texts, presents a close competition among the models. This implies that the inclusion of both content words and their ordering are well-represented across models.

\subsection{MRR Comparison}

An additional analysis conducted to assess the consistency between reward models and metrics used for evaluating the quality of the generated responses. This analysis involved the utilization of manual annotations for answers corresponding to 100 random questions.

In this analysis, MRR calculated for varying values of $k$, where $k$ denotes the top-ranked answers accordingly to some metric. The Mean Reciprocal Rank at k (MRR@k) is a statistical measure for evaluating any process that produces a list of possible responses to a sample of queries, ordered by probability of correctness. If we have $A$ answers, and $R_i$ is the rank of the first relevant document for query $i$ (considering only the top $k$ documents), then the MRR@k is:

\begin{equation}
\small
MRR@k = \frac{1}{A} \sum_{i=1}^{A} \left(\frac{1}{R_i}\right) \text{ if } R_i \leq k \text{; else } 0
\end{equation}

This method allows for understanding how effectively the different metrics rank the correct answers among its top predictions.

Table~\ref{table:mrr_comparison} presents the MRR@10 scores, indicating the MRR values when considering the top 10 ranked samples. Notably, Rouge 2 and Rouge 1 metrics exhibit higher values, which implies that they are key metrics in assessing the accuracy of the generated responses. However, the trained reward models display superior performance compared to both the SacreBLEU and BertScore metrics.

\begin{table*}
\centering
\begin{tabular}{l@{\hspace{3mm}}c@{\hspace{3mm}}c@{\hspace{3mm}}c}
\hline
 & \textbf{Base 125M} & \textbf{SFT 125M} & \textbf{RLHF 125M} \\
\hline
\textbf{SacreBLEU} & 0.4107 & 0.3709 & 0.3262 \\
\textbf{Rouge 1} & \textbf{0.4792} & \textbf{0.4532} & 0.4091 \\
\textbf{Rouge 2} & 0.4011 & 0.4453 & 0.4220 \\
\textbf{BertScore} & 0.2913 & 0.3403 & 0.3300 \\
\textbf{Reg. Reward} & 0.4015 & 0.3867 & \textbf{0.4296} \\
\textbf{Contr. Reward} & 0.4302 & 0.3787 & 0.3527 \\
\hline
\end{tabular}
\caption{\label{table:mrr_comparison}
Comparison of MRR@10 scores for different models and metrics. The values represent the MRR scores considering the top 10 ranked samples.
}
\end{table*}

\subsection{Correlation Analysis}

In addition to the previous evaluations, a correlation analysis carried out among the assessment methods utilized. Specifically, the Spearman correlation coefficient was computed to understand the relationships between the various metrics. The Spearman correlation coefficient is a nonparametric measure that evaluates the strength and direction of the association between two ranked variables.

Appendix~\ref{sec:corr_coef_appendix} contains tables that compare the cross-correlation coefficients of the metrics for each model generations. Upon examination, a prominent correlation between the rankings of Rouge 1 and Rouge 2 is evident. Furthermore, the reward regression model exhibits a moderate correlation when responses are generated by the fine-tuned models. Interestingly, BertScore demonstrates little to no correlation, or even a negative correlation, with the other metrics. This raises questions about its reliability as a comparative measure in this context. Additionally, it is notable that the reward models display minimal correlation amongst themselves, when trained through different methodologies.

\section{Discussion}

The study focused on the generating QA highlights the effective implementation of the RLHF in the intricate domain. This method outperforms SFT technique, marking its superiority in terms of metrics performance. Moreover, the application of RLHF has demonstrated that it's possible to competitively train smaller models, showcasing its efficacy even in scenarios with limited resources.

Regarding the scoring parameters, the study draws attention to the utility of Rouge scores in gauging response precision. This implies a potential edge of Rouge over alternative scoring systems like SacreBLEU and BertScore in certain contexts.

However, there exists ambiguity in the MRR results for BertScore and SacreBLEU metrics when juxtaposed with the outcomes from the trained reward models. This raises questions about the adequacy of these metrics for the programming domain, which is hallmarked by complex semantic relationships and a plethora of correct answers. This ambiguity is further cemented by near-zero Spearman correlations associated with various linguistic metrics.

These findings not only provide a deeper understanding of RLHF's potential and boundaries but also emphasize the necessity for diverse, domain-specific methods when evaluating generation quality. In this context, the programming domain serves as an exemplar. This research's insights could stimulate further advancements in the development of novel and more suitable metrics for similar complex domains.

\section{Conclusion}

In conclusion, our study has demonstrated the effectiveness of RLHF in enhancing the performance of small LLMs like GPT Neo 125M in the programming domain. Our experiments focused on fine-tuning the model using user-generated responses from Stack Overflow, employing two reward model training strategies, regression scores and contrastive scores, with PPO.

The study also highlights the critical role of employing the right evaluation measures. While Rouge scores effectively captured response accuracy, other metrics like BertScore and SacreBLEU presented ambiguities, especially when juxtaposed with the results from the trained reward models. This disparity, brought into sharper focus by near-zero Spearman correlations, implies that traditional metrics might not suffice for complex fields such as programming. These domains are marked by intricate semantic relationships and a broad spectrum of valid answers.

As we look to the future, we envision testing our methodologies and experiment setups on larger models to assess the scalability of our approach and verify the consistency of our results. We anticipate that these further investigations would provide valuable insights into the behavior and performance of these larger models under RLHF based fine-tuning, thereby expanding the scope of our current study.

The insights derived from our research enrich our understanding of both the potential and the limitations of RLHF. They also underline the necessity for tailored evaluation methods in complex domains. As we persist in honing and formulating innovative techniques for efficient generation, the lessons gleaned from our work will undoubtedly prove invaluable.

\section*{Limitations}
In this study, we attempted to investigate the application of the GPT Neo model with 125M parameters in assessing the quality of linguistic metrics in the Usage API subcategory of question-and-answer data. We acknowledge several limitations that need to be taken into account when interpreting the results.

First, the data used in the experiments is domain-specific, sourced exclusively from the Usage API subcategory, which lacks code blocks. Although our findings demonstrate discrepancies between linguistic metrics within this chosen domain, their generalizability to other question-and-answer categories remains unclear. It's possible that our reward model may perform differently when applied to more diverse datasets with varied question types and content, including those that incorporate code blocks.

Second, the application of the small GPT Neo model with 125M parameters represents a significant limitation in terms of both computational capacity and the model's semantic understanding. The constraints of our computing resources, specifically the usage of 2 Nvidia A6000 GPUs and the necessity to accommodate three models during the RLHF training, have imposed certain restrictions. Owing to VRAM limitations, portions of the question context were omitted during training, potentially undermining the model's ability to fully grasp the semantic relations in the language.

Another caveat concerns the scale of the model. While our experiments illustrated the small model's ability to enhance results to levels comparable to its larger counterparts, the behavior of the larger models under similar experimental conditions is yet to be understood. This question remains open and warrants further investigation in future research.

In summary, our study provides valuable insights into the use of smaller GPT Neo models for assessing linguistic metrics, but the highlighted limitations underscore the need for additional research in broader data contexts, with larger models, and considering the intricate facets of language translation and reformulation.

\bibliography{paper}
\bibliographystyle{acl_natbib}

\appendix

\setcounter{figure}{0}
\renewcommand{\thefigure}{A\arabic{figure}}

\section{Spearman correlation tables}
    
\label{sec:corr_coef_appendix}
\begin{figure}[h]
\centering
\includegraphics[width=0.35\textwidth]{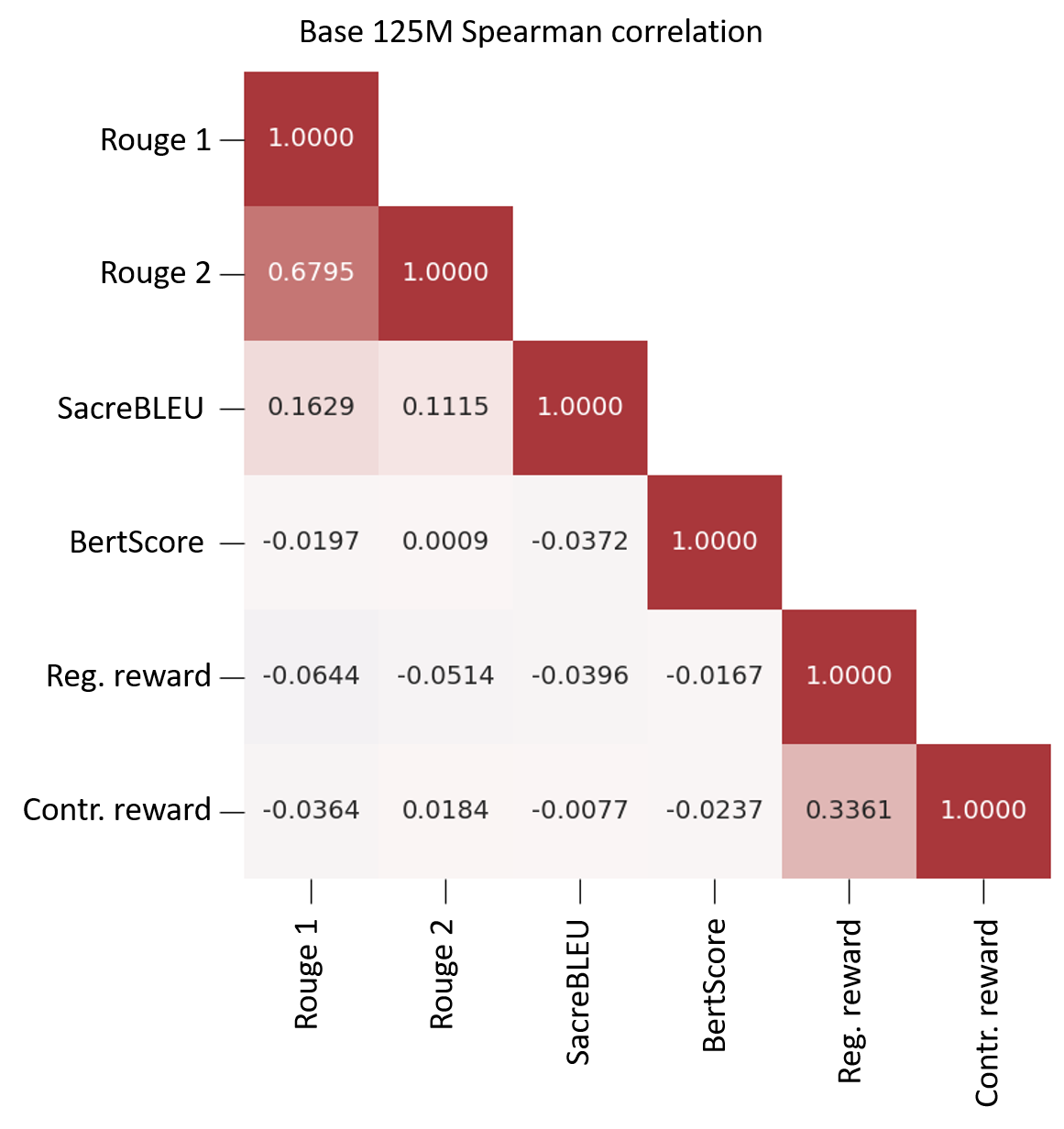}
\caption{\label{fig:base_corr_coef}Spearman correlation coefficients for Base model}
\end{figure}

In this appendix, we present \cref{fig:base_corr_coef,fig:sft_corr_coef,fig:rlhf_corr_coef} that feature comparative tables of Spearman's correlation coefficients for several evaluation metrics: Rouge 1, Rouge 2, SacreBLEU, and BertScore and used two variations of reward models, the regressive and contrastive. They based on the generations produced by three distinct models. These models are the Base GPT Neo 125M, the SFT GPT Neo 125M, and the RLHF GPT Neo 125M, respectively.

\begin{figure}[h]
\label{fig:sft_corr_coef}
\centering
\includegraphics[width=0.35\textwidth]{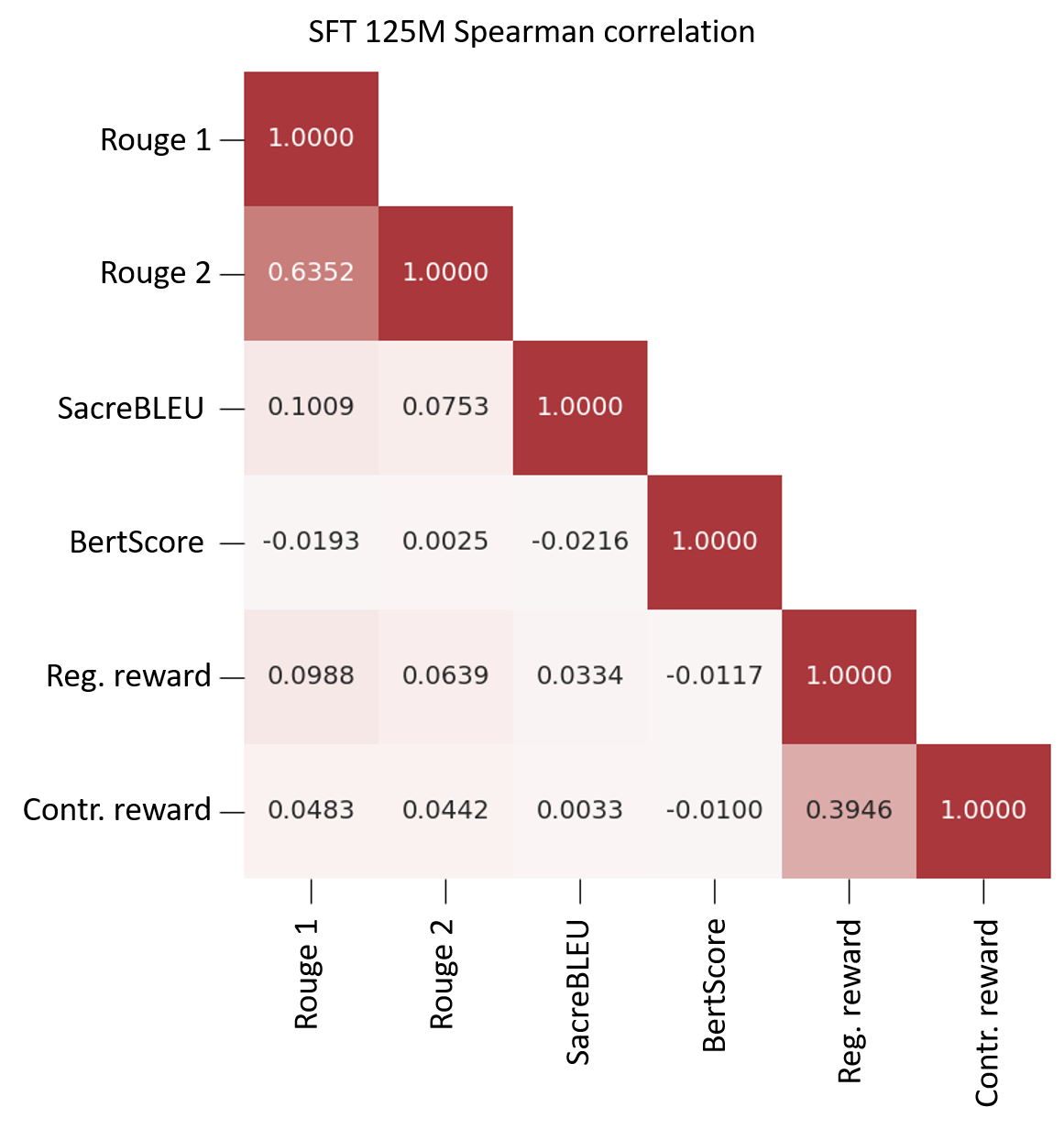}
\caption{\label{fig:sft_corr_coef}Spearman correlation coefficients for SFT model}
\end{figure}

\begin{figure}[h]
\label{fig:rlhf_corr_coef}
\centering
\includegraphics[width=0.35\textwidth]{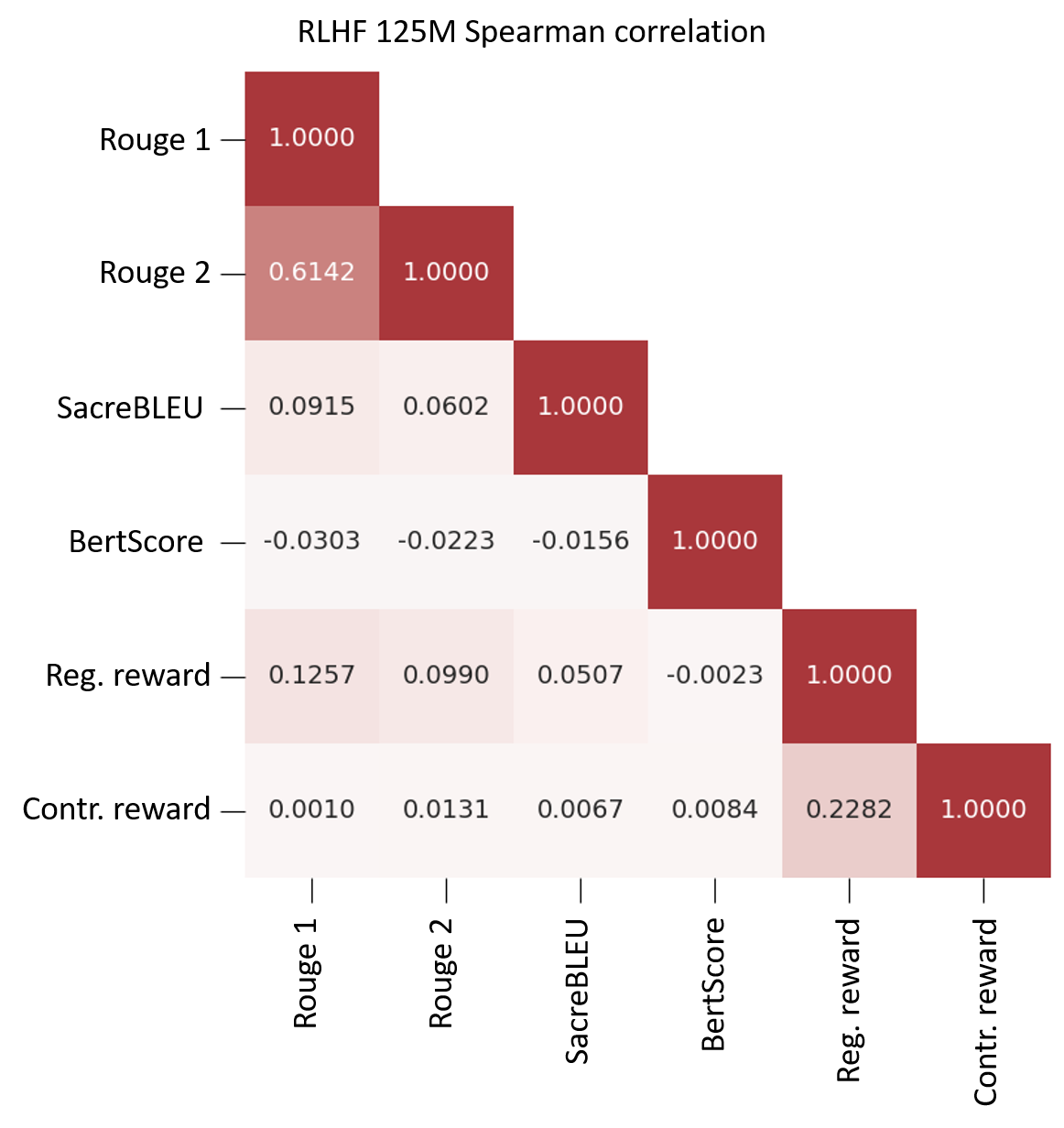}
\caption{\label{fig:rlhf_corr_coef}Spearman correlation coefficients for RLHF model}
\end{figure}

\end{document}